\documentclass[runningheads]{llncs}
\usepackage[T1]{fontenc}
\usepackage{graphicx}
\usepackage{xcolor}
\usepackage{amsmath}
\usepackage{booktabs} 
\definecolor{Gray}{gray}{0.9} 

\usepackage{amssymb}
\usepackage{pifont}
\newcommand{\cmark}{\text{\ding{51}}}%
\newcommand{\xmark}{\text{\ding{55}}}%
\usepackage{multirow} 
\usepackage[misc,geometry]{ifsym}
\usepackage[pagebackref,breaklinks,colorlinks]{hyperref}

\usepackage{amsmath,amsfonts,bm}

\usepackage{bbm}
\usepackage{bm}

\usepackage{booktabs}       
\usepackage{amsfonts}       
\usepackage{nicefrac}       
\usepackage{microtype}      
\usepackage{color}
\usepackage{array}
\usepackage{multirow}
\usepackage{dirtytalk}
\usepackage[para,online,flushleft]{threeparttable}
\usepackage{colortbl}
\usepackage[normalem]{ulem}

\definecolor{Crimson}{rgb}{0.86, 0.08, 0.24}
\definecolor{DarkGreen}{rgb}{0.10, 0.55, 0.10}
\definecolor{RoyalBlue}{rgb}{0.20, 0.60, 0.86}
\definecolor{DarkCyan}{rgb}{0.0, 0.54, 0.54}
\definecolor{Gray}{gray}{0.9}
\definecolor{ChromeYellow}{rgb}{1.0, 0.65, 0.0}
\definecolor{Salmon}{rgb}{0.98, 0.50, 0.45}
\definecolor{LightGreen}{rgb}{0.93,0.98,0.96}
\ifx \submission \undefined

\else

\fi

\usepackage[most]{tcolorbox}
\newtcolorbox{prompt}{
    colback=gray!10,
    colframe=gray!40,
    fonttitle=\bfseries,
    coltitle=black,
    colbacktitle=gray!40,
    enhanced,
    drop shadow=black!5!white,
    left=8mm,
    right=8mm,
    top=3mm,
    bottom=3mm,
    boxsep=1mm,
    title=Text Prompt:
    }
\newtcolorbox{fs_template}{
    colback=gray!10,
    colframe=gray!40,
    fonttitle=\bfseries,
    coltitle=black,
    colbacktitle=gray!40,
    enhanced,
    drop shadow=black!5!white,
    left=8mm,
    right=8mm,
    top=3mm,
    bottom=3mm,
    boxsep=1mm,
    title=Few-Shot Template:
    }
\newtcolorbox{fs_query}{
    colback=gray!10,
    colframe=gray!40,
    fonttitle=\bfseries,
    coltitle=black,
    colbacktitle=gray!40,
    enhanced,
    drop shadow=black!5!white,
    left=8mm,
    right=8mm,
    top=3mm,
    bottom=3mm,
    boxsep=1mm,
    title=Few-Shot Queries:
    }
\newtcolorbox{bkg_template}{
    colback=gray!10,
    colframe=gray!40,
    fonttitle=\bfseries,
    coltitle=black,
    colbacktitle=gray!40,
    enhanced,
    drop shadow=black!5!white,
    left=8mm,
    right=8mm,
    top=3mm,
    bottom=3mm,
    boxsep=1mm,
    title=Clinical Report Summarizer Template (to GPT-4):
    }
\newtcolorbox{bkg_query}{
    colback=gray!10,
    colframe=gray!40,
    fonttitle=\bfseries,
    coltitle=black,
    colbacktitle=gray!40,
    enhanced,
    drop shadow=black!5!white,
    left=8mm,
    right=8mm,
    top=3mm,
    bottom=3mm,
    boxsep=1mm,
    title=Few-Shot Background Queries:
    }
\newtcolorbox{bkg_prompt}{
    colback=gray!10,
    colframe=gray!40,
    fonttitle=\bfseries,
    coltitle=black,
    colbacktitle=gray!40,
    enhanced,
    drop shadow=black!5!white,
    left=8mm,
    right=8mm,
    top=3mm,
    bottom=3mm,
    boxsep=1mm,
    title=Retrieved Background:
    }

\begin{document}
\title{Insight: A Multi-Modal Diagnostic Pipeline using LLMs for Ocular Surface Disease Diagnosis}
\titlerunning{A Multi-Modal Diagnostic Pipeline using LLMs for OSD Diagnosis}
%
\author{Chun-Hsiao Yeh\inst{1,2} \and
Jiayun Wang\inst{1,3} \and
Andrew D. Graham\inst{1,2} \and
Andrea J. Liu\inst{1} \and
Bo Tan\inst{1} \and
Yubei Chen\inst{4} \and
Yi Ma\inst{2,5} \and
Meng C. Lin\inst{1,2}\textsuperscript{(\Letter)}}
%

\authorrunning{CH. Yeh et al.}

%
\institute{
Clinical Research Center, University of California, Berkeley, Berkeley, CA, USA
\and University of California, Berkeley, Berkeley, CA, USA \\
\email{\{daniel\_yeh, mlin\}@berkeley.edu} \\
\and California Institute of Technology, Pasadena, CA, USA
\and University of California, Davis, Davis, CA, USA
\and University of Hong Kong, Hong Kong SAR, China
}
\maketitle              
\begin{abstract}
Accurate diagnosis of ocular surface diseases is critical in optometry and ophthalmology, which hinge on integrating clinical data sources (e.g., meibography imaging and clinical metadata). Traditional human assessments lack precision in quantifying clinical observations, while current machine-based methods often treat diagnoses as multi-class classification problems, limiting the diagnoses to a predeﬁned closed-set of curated answers without reasoning the clinical relevance of each variable to the diagnosis. To tackle these challenges, we introduce an innovative multi-modal diagnostic pipeline (MDPipe) by employing large language models (LLMs) for ocular surface disease diagnosis. 
We first employ a visual translator to interpret meibography images by converting them into quantifiable morphology data, facilitating their integration with clinical metadata and enabling the communication of nuanced medical insight to LLMs. To further advance this communication, we introduce a LLM-based summarizer to contextualize the insight from the combined morphology and clinical metadata, and generate clinical report summaries. Finally, we refine the LLMs' reasoning ability with domain-specific insight from real-life clinician diagnoses. Our evaluation across diverse ocular surface disease diagnosis benchmarks demonstrates that MDPipe outperforms existing standards, including GPT-4, and provides clinically sound rationales for diagnoses. The project is available at \url{https://danielchyeh.github.io/MDPipe/}.

\keywords{Multimodality \and Large Language Models \and Ocular Surface Disease Diagnosis}
\end{abstract}
\section{Introduction}
 Ocular surface diseases (OSD) are various disorders impacting the anterior segment of the eye, with Dry Eye (DE) being the most prevalent. DE significantly impacts ocular surface health, vision, and quality of life, and is a predominant reason for eye care visits globally~\cite{craig2017tfos,rouen2018dry}. Evaporative DE, the most common subtype, is primarily attributed to Meibomian Gland Dysfunction (MGD)~\cite{nichols2011international} which is characterized by the glands' failure to secrete a sufficiently thick and well-organized lipid layer~\cite{bron2004functional}. This results in increased tear evaporation, tear film thinning and destabilization, leading to tear hyperosmolarity and premature tear breakup, culminating in the discomfort associated with DE~\cite{teo2020meibomian}. Traditional diagnosis involves clinical assessments such as measurement of fluorescein tear breakup time (FTBUT), noninvasive keratograph breakup time (NIKBUT), grading of MG expressate, meibography imaging, as well as administration of symptom instruments such as the Ocular Surface Disease Index (OSDI)~\cite{pult2011relationship}. Yet, these approaches are time-consuming~\cite{wang2021quantifying} and have poor reproducibility~\cite{nichols2004repeatability}, particularly in quantifying meibomian gland (MG) atrophy~\cite{finis2015evaluation,arita2009proposed} where estimations are subject to human bias and lack the accuracy for a definitive diagnosis.

\begin{figure}[t!]
\includegraphics[width=\textwidth]{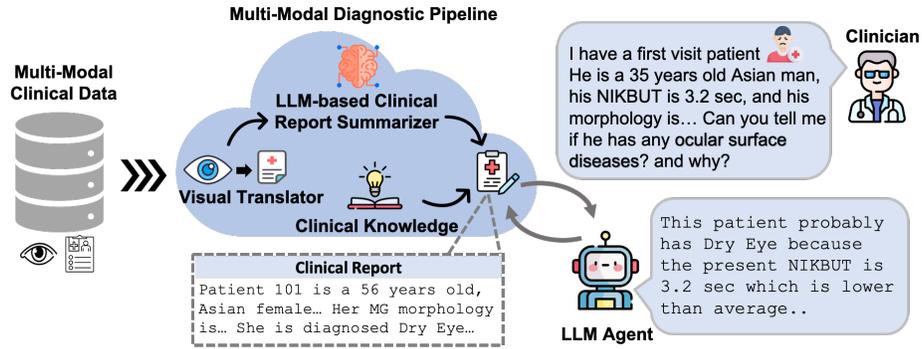}
\caption{Multi-modal diagnostic pipeline using LLMs for OSD diagnosis. The proposed pipeline utilizes 1) a visual translator to
transform meibography images into quantifiable MG morphology, 2) an LLM-based summarizer to craft clinical reports, and 3) the integration of clinical knowledge to augment LLM's capability in diagnosing OSD.} \label{fig-teaser}
\end{figure}

Recently, efforts to employ machine learning (ML) in the diagnosis of OSD have seen notable advancements. Initial efforts have concentrated on quantifying MG morphology through imaging techniques~\cite{saha2022automated,yeh2021meibography} and segmentation models for MG feature extraction~\cite{wang2019deep,prabhu2020deep,wang2021quantifying}. As research evolves, there is a growing emphasis on combining visual assessments of the MGs with clinical metadata~\cite{lin2023lifestyle}. However, incorporating a diverse set of clinical variables into predictive classification models results in closed-set predictions based on predefined sets of answers, and treats clinical variables merely as data points without semantic relationships or diagnostic rationale. Thus, understanding the clinical implications of the metadata and how these variables relate to conditions such as DE remains an ongoing challenge that requires further exploration.

The emergence of Multimodal Large Language Models (MLLMs)~\cite{openai2023gpt4,zhu2023minigpt,liu2023visual} brings new light to this problem. MLLMs leverage the powerful abilities of LLMs~\cite{openai2023gpt,touvron2023llama} while integrating visual data. MLLMs, when utilized for diagnostic purposes~\cite{li2024llava,liu2023m,zhang2023biomedgpt}, are capable of producing hundreds to thousands of free-form answers with clinical reasoning, instead of being limited to closed-set predictions. However, several critical analyses of MLLMs reveal persistent challenges in accurately representing visual data~\cite{tong2024eyes,zhai2024investigating}. This issue becomes particularly pronounced when dealing with rare domains or complex imagery, such as meibography (see Fig.~\ref{fig-translator}).

Addressing this gap, we pose two questions: 1) Can a model process meibography images with the same level of attention and detail as a human clinician? and 2) Can the model make a precise and accurate diagnosis by focusing on specific MG morphological features and quantifying those observations along with clinical metadata? To answer these questions, we propose a novel multi-modal diagnostic pipeline (MDPipe) that integrates patient clinical metadata and
visual data for OSD diagnosis through the use of LLMs (see Fig.~\ref{fig-teaser}). Specifically, we contribute in three major aspects:

\begin{itemize}

    \item We propose a visual translator to convert meibography images into quantifiable MG morphology data through an instance segmentation network. This transformation allows us to bridge the visual data with clinical metadata, effectively marrying detailed morphological insights with clinical context.
    \item To advance this integration, we introduce an LLM-based summarizer to generate clinical report summaries to contextualize the insights from both the non-narrative clinical metadata and the MG morphology data.
    \item We further collect clinical knowledge using real-life clinician diagnoses to refine the LLMs with nuanced, domain-specific knowledge.

\end{itemize}

\section{Method}

\subsection{Problem Formulation}
Our objective is to develop a multi-modal pipeline integrating patient clinical metadata \( \mathbf{M} = (\mathbf{C}, \mathbf{D}) \) where \( \mathbf{C} \) represents clinical measurements and \( \mathbf{D} \) denotes the clinician's diagnosis, with meibography images \( \mathbf{I} \). This integration aims to use LLMs to enhance ocular disease diagnostics by fine-tuning an LLM \(p_{\theta}\) to diagnose patients using summarized clinical data. 

We first use a report summarizer \(\mathcal{S}\) implemented using GPT-4~\cite{openai2023gpt4} to convert the metadata \(\mathbf{M}\) and image data \(\mathbf{I}\) into question-answer pairs \((\mathbf{Q}, \mathbf{A}) = \mathcal{S}(\mathbf{M}, \mathcal{V}(\mathbf{I}))\), where \(\mathcal{V}\) is a visual translator (in practice, implemented using a pre-trained segmentation model) that extracts MG morphology from an image \( \mathbf{I} \). The ultimate goal is to refine the LLM parameters \( \theta^* \) through the maximization of the conditional log-likelihood, formalized as: \(\theta^* = \mathrm{arg\,max}_\theta \sum_{i=1}^{N} \log p_{\theta}(\mathbf{A}_i | \mathbf{Q}_i)\), where \( N \) is the total number of cases. In practice, given new clinical data \( \mathbf{C^*} \) and image \( \mathbf{I^*} \), we can generate a new query \( \mathbf{Q}^* = \mathcal{S}(\mathbf{C^*}, \mathcal{V}(\mathbf{I^*})) \), which can then be fed into the LLMs to predict the corresponding diagnosis \( \mathbf{A^*} \).

\begin{figure}[t!]
\includegraphics[width=\textwidth]{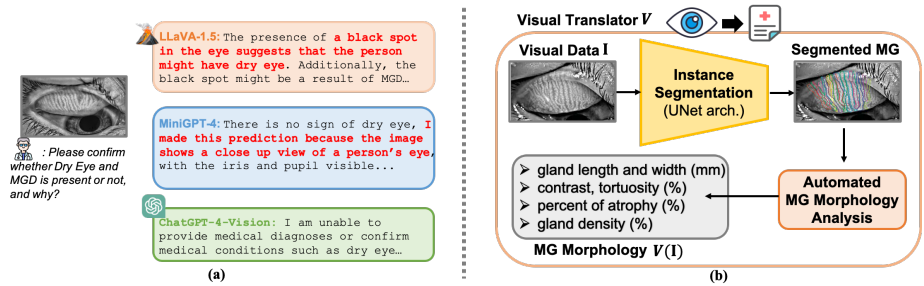}
\caption{(a) Illustration of the limitations of current MLLMs in processing visual data, including: 1) producing vague interpretations; 2) not delivering clinical significance for ocular surface diseases, such as labeling "a black spot in the eye". (b) Our visual translator \(\mathcal{V}\) is designed to interpret visual data \( \mathbf{I} \) by converting them into quantifiable MG morphology data.} \label{fig-translator}
\end{figure}

\subsection{Visual Translator}
MLLMs have known limitations when processing visual data and generating corresponding representations. The intricate details captured in meibography images present a particular challenge (see Fig.~\ref{fig-translator}-(a)) due to their complexity and the nuanced information they hold. To address this, we set out to construct a model to process meibography images with the same level of attention and detail as a human clinician and to quantify the observations. Building on prior works~\cite{de2017semantic,wang2021quantifying}, we introduced a visual translator \( \mathcal{V} \), designed to convert meibography images \( \mathbf{I} \) into quantifiable MG morphology data. This transformation served as a bridge, linking the raw visual data with clinical metadata and thus weaving together rich morphological detail with pertinent clinical context.

The visual translator operated through a two-step process. (see Fig.~\ref{fig-translator}-(b)) Firstly, meibography images were passed to an instance segmentation network~\cite{de2017semantic} that automatically delineated individual gland regions. Secondly, a detailed quantification of morphological features at the MG level was performed. By automating these steps, the model enhanced the observational capabilities of clinicians, providing a quantitative assessment of the glandular characteristics that were emerging as important factors in diagnosing OSD.

With the visual translator, we were able to precisely measure morphological features such as percent atrophy, gland density, average gland local contrast, and dimensions of gland length, width, and tortuosity~\cite{yeh2019repeatability}. These measures were quantified, allowing for an accurate depiction of the MG morphology, beyond what manual estimation could achieve.

\subsection{LLM-based Clinical Report Summarizer}
As illustrated in Fig.~\ref{fig-llm_summarizer} (left), the clinical metadata and MG morphology, derived via the visual translator, presented in a discrete and non-narrative form. Our goal was to input this fragmented data into GPT-4~\cite{openai2023gpt4} to synthesize a cohesive clinical report summary for each case. The prompting structure provided to GPT-4, depicted in Fig.~\ref{fig-llm_summarizer} (middle), had three main components: 1)  a task description, 2) supporting examples, and 3) prompting the clinical metadata. Each component was designed to guide GPT-4 toward generating an informative and clinically relevant summary, which we delineated as follows:

\begin{figure}[t!]
\includegraphics[width=\textwidth]{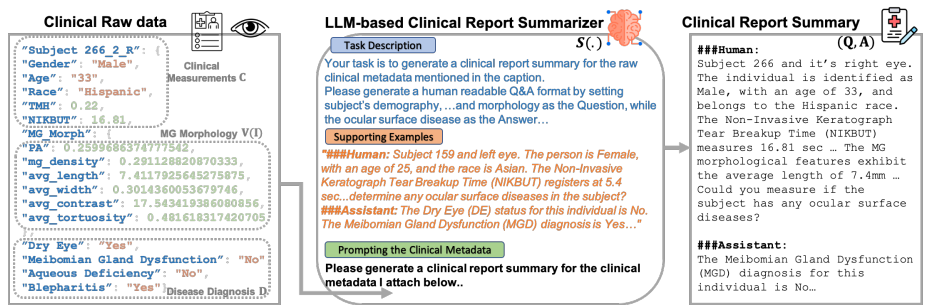}
\caption{We employed an LLM-based summarizer to generate Q\&A clinical reports to contextualize insights from both the non-narrative clinical metadata and MG morphology to enhance LLMs' learning capability.} 
\label{fig-llm_summarizer}
\end{figure}

\subsubsection{Task Description.} 
We first directed the model to perform the task of generating a clinical report summary from clinical raw data. The instruction specified the Q\&A format, with the subject's demographics, clinical, and morphological data forming the questions, and the suspected OSD as the answer. This concise directive ensured that the output was clinically informative and formatted.

\subsubsection{Supporting Examples.}
Drawing  from~\cite{lian2023llm}, we enhanced the model's understanding by supplementing the task description with manually curated examples. These examples served to illustrate the desired approach for crafting clinical report summaries and were strategically employed to resolve ambiguities, guiding GPT-4 toward producing outputs that aligned with clinical reporting standards.

\subsubsection{Prompting the Clinical Metadata.}
Upon presenting the supporting examples, we proceeded to prompt the model with a request to generate a clinical report summary, attaching the raw clinical metadata subsequently. This approach enabled the model to synthesize a clinical report summary that adhered closely to our predefined instructions. Template can be found in the Appendix.

\subsection{Refinement of LLMs with Clinical Knowledge}

\subsubsection{Fundamental Clinical Knowledge.}
To refine an LLM for ocular disease diagnosis, we enhanced it with targeted knowledge using DE-specific clinical terms and definitions extracted from 15 published DE clinical trials conducted between 2015 and 2023~\cite{tauber2021randomized,schmidl2020influence}. This information was formatted into Q\&A pairs with GPT-4, emphasizing common trial criteria like specific FTBUT ranges and OSDI score thresholds. Our fine-tuning process re-focused the LLM on key medical data from DE clinical trials, potentially already within its training set, to deepen its clinical understanding.

\subsubsection{Real-Life Clinician Diagnoses.}
For the second refinement phase, we integrated authentic clinical diagnostic cases to ground the LLMs in empirical medical expertise. A dataset comprising 20 real-world cases was compiled by human clinicians, encompassing both the clinicians' diagnosis of various OSD (e.g., MGD, blepharitis), and the corresponding rationale underpinning these clinical judgments. This dataset was effectively transformed into a series of Q\&A pairs, effectively inculcating the LLMs with the nuanced reasoning and decision-making processes typical of seasoned clinicians.

\section{Experimental Setup}

\subsubsection{Datasets.}

We utilized two multimodal datasets: CRC~\cite{wang2019deep,wang2021quantifying} and DREAM~\cite{asbell2018dry,hussain2020dry}, which included meibography images and clinical metadata. CRC is a curated dataset that spans a spectrum from normal to moderate DE conditions. Conversely, the DREAM dataset primarily contains data from patients with moderate to severe DE conditions. We merged these two datasets and, via our processing pipeline, converted them into a QA-based clinical report format conducive to diagnostic tasks. The resultant composite dataset contained a total of 3513 entries. Train/test split is 90\%/10\%. Training set has 1903 metadata-only and 1257 image+metadata instances; Test set has 198 metadata-only and 155 image+metadata instances. There are a total of 878 subjects.

\subsubsection{Metric.} 
For quantitative analysis, we followed established benchmarks~\cite{wu2023pmc}, utilizing accuracy and F1 score. We also incorporated sensitivity (SN) and specificity (SP) metrics. On the qualitative front, we engaged human clinicians in a user study to grade the LLM's diagnostic outputs across multiple dimensions.

\subsubsection{Implementation Details.}
We chose the 7B and 13B versions of pre-trained LLaMA-2~\cite{touvron2023llama}. We obtained the model checkpoints from the official Huggingface ({\texttt{NousResearch/Llama-2-7b-chat-hf, NousResearch/Llama-2-13b-chat-hf}}). All models were able to train on four NVIDIA Geforce RTX 3090 GPUs (average training time: $\sim$8 hrs). We used the AdamW optimizer with 0.03 warm-up ratio and a learning rate of 2e-4. More details can be found in the Appendix.

\section{Results}

\begin{table}[t!]
\caption{Comparison between general and medical domain-tuned LLMs for diagnosing ocular diseases: Dry Eye (DE), Meibomian Gland Dysfunction (MGD), and Blepharitis. Evaluation criteria include accuracy, sensitivity (SN), specificity (SP), and F1 score. Our MDPipe outperforms existing benchmarks across three ocular diseases.}
\label{tab:comparison}
\centering
\begin{tabular}{l|cccc|cccc|cccc} 
\toprule
\multirow{2}{*}{Method / Disease} & \multicolumn{4}{c|}{\textbf{DE}} & \multicolumn{4}{c|}{\textbf{MGD}} & \multicolumn{4}{c}{\textbf{Blepharitis}}  \\
   & Acc. & SN & SP & F1 & Acc. & SN & SP & F1 & Acc. & SN & SP & F1 \\
\midrule
\multicolumn{13}{l}{\textit{General LLMs without fine-tuning}} \\
Llama~\cite{touvron2023llama} & 49.8 & 93.2 & 14.7 & 60.5 & 40.6 & 88.7 & 17.1 & 55.9 & 44.7 & 28.5 & 55.3 & 30.8 \\ 
GPT-3.5~\cite{openai2023gpt} & 57.7 & 86.7 & 32.7 & 64.9 & 48.6 & 95.5 & 25.6 & 60.6 & 49.2 & 31.3 & 61.9 & 33.8 \\ 
Llama2-7B~\cite{touvron2023llama} & 63.9 & 88.2 & 38.6 & 66.6 & 52.7 & 83.2 & 23.3 & 62.3 & 47.4 & 31.8 & 59.3 & 34.4 \\ 
GPT-4~\cite{openai2023gpt4} & 70.7 & 77.1 & 66.3 & 67.7 & 65.2 & 65.7 & 76.8 & 65.5 & 58.2 & 39.3 & 72.9 & 48.8 \\ 
\midrule
\multicolumn{13}{l}{\textit{LLMs fine-tuned on medical domain data}} \\  
Med-Alpaca~\cite{han2023medalpaca} & 62.5 & 87.3 & 33.5 & 70.3 & 53.4 & 84.7 & 28.2 & 61.9 & 54.9 & 53.8 & 55.8 & 49.7 \\
PMC-LLaMA~\cite{wu2023pmc} & 73.3 & 73.1 & 77.7 & 75.8 & 63.6 & 70.7 & 61.5 & 64.7 & 60.5 & 50.3 & 74.4 & 56.8 \\
\midrule
MDPipe-7B (ours) & 86.9 & 89.3 & 84.3 & 87.8 & \textbf{76.1} & \textbf{67.2} & \textbf{81.7} & \textbf{69.2} & 71.2 & 56.3 & 79.7 & 63.8 \\
MDPipe-13B (ours) & \textbf{89.5} & \textbf{88.2} & \textbf{91.0} & \textbf{89.9} & 74.4 & 61.4 & 82.9 & 65.7 & \textbf{73.1} & \textbf{58.7} & \textbf{80.1} & \textbf{65.1} \\
\bottomrule
\end{tabular}  
\end{table}

\subsubsection{Comparison of General and Medical LLMs.}
In Table~\ref{tab:comparison}, we evaluate the performance of various LLMs in diagnosing ocular diseases, specifically Dry Eye (DE), Meibomian Gland Dysfunction (MGD), and Blepharitis. Notably, GPT-4 demonstrates consistent performance across all three ocular diseases. This proficiency is likely attributed to its extensive pre-training across a broad dataset, encompassing a range of medical knowledge, which may contribute to its enhanced ability to generalize across different domains.

When considering models fine-tuned on medical domain data, PMC-LLaMA showed notable improvements over general LLMs, achieving an accuracy of 73.3\% for DE, demonstrating the advantage of domain-specific fine-tuning. However, our MDPipe models significantly outperform all other models, achieving an accuracy of 86.9\% for DE, 81.7\% for MGD, and 79.7\% for Blepharitis.

\subsubsection{Ablation Studies.} In Table~\ref{tab:ablation}, we investigated the impact of various training variables within our MDPipe on the diagnostic accuracy of the pre-trained LLaMA2-7B model. The variables considered were clinical metadata, MG morphology data, MG expression (quality and quantity scores), and real clinician diagnoses. We found the inclusion of MG morphology notably improved performance, especially in MGD diagnosis, increasing accuracy from 65.5\% to 74.4\%. This suggests the significant role of morphological features in detecting MGD.

\begin{table}[t!]
\centering
\caption{The impact of various training variables within our MDPipe on ocular disease diagnosis. Notably, it is observed that MG morphology is essential in MGD diagnosis.}
\label{tab:ablation}
\begin{tabular}{lcccc|ccc} 
\toprule
\multirow{2}{*}{Pretrain} & \multicolumn{4}{c|}{+ Training Variables in MDPipe} & \multicolumn{3}{c}{Diagnosis Acc. (\%)} \\
     &Metadata |& Morphology |& MG-Express. |& Real Diag. & \textbf{DE} & \textbf{MGD} & \textbf{Bleph.} \\
\midrule
\multirow{4}{*}{LLaMA2}  & \cmark & \xmark & \xmark & \xmark & 83.5 & 65.5 & 69.4 \\
& \cmark & \cmark & \xmark & \xmark & 84.1 & 74.4 & 68.8 \\
& \cmark & \cmark & \cmark & \xmark & 85.8 & 75.6 & 70.1 \\
& \cmark & \cmark & \cmark & \cmark & \textbf{86.9} & \textbf{76.1} & \textbf{71.2} \\
\bottomrule
\end{tabular}
\end{table}

\begin{figure}[t!]
\includegraphics[width=\textwidth]{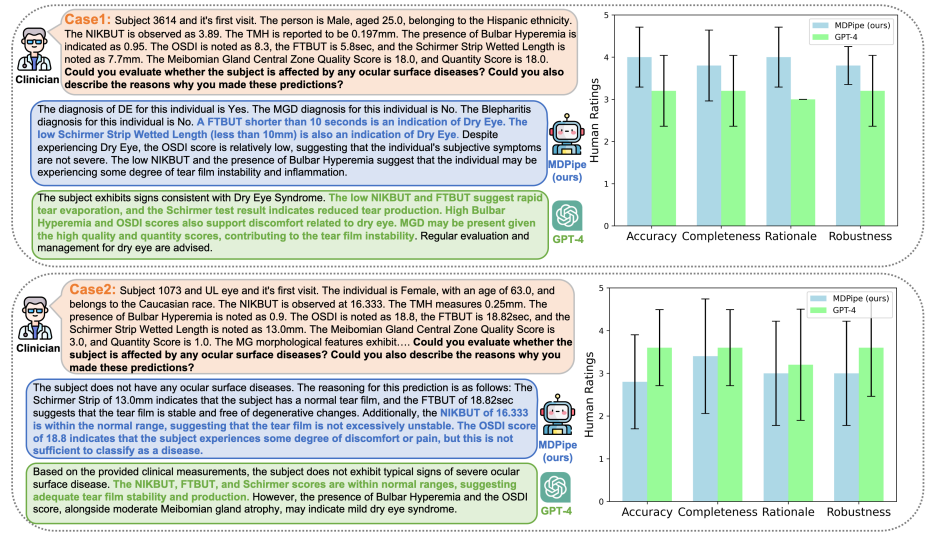}
\caption{Comparative evaluation and clinician study between MDPipe and GPT-4.}

\label{fig-compare}
\end{figure}

\subsubsection{User (Clinician) Preference Study.}
A  user (clinician) preference study was performed to evaluate the performance of our MDPipe and GPT-4 in diagnosing OSD. A series of three random cases were presented with the appropriate diagnoses and rationale. Specifically, 5 clinicians were masked as to which model produced each output, and then asked to read and rate the two models' output on a scale from 1 (poor) to 5 (best) regarding 1) clinical accuracy, 2) diagnostic completeness, 3) diagnostic rationale, and 4) the model's robustness to handle ambiguous or incomplete patient data. 

Figure~\ref{fig-compare} depicts the comparative study between MDPipe and GPT-4 using two clinical cases. In case 1, our MDPipe surpasses GPT-4, as evidenced by higher scores in accuracy and rationale. MDPipe's diagnostic responses apply DE criteria (e.g., FTBUT < 10 sec and Schirmer's test measurements < 10mm), which suggests a successful incorporation of DE clinical trials into the LLMs as detailed in Section 2.4. Conversely, GPT-4 seems to lack a nuanced interpretation of clinical measurements. For instance, one clinician remarks \textit{"The OSDI is <13 and normal, I would not consider 0.95 bulbar hyperemia to be signs of dry eye"}, implying a potential gap in GPT-4's understanding of the subtleties within clinical scale values and their association with disease states. In case 2, MDPipe exhibits shortcomings in  completeness, as noted by a clinician's feedback, \textit{"The diagnosis does not take into account MG expression and morphological data..."} pointing out the omission of some variables, indicating a potential oversight of comprehensive variable consideration in the diagnostic process.

\section{Conclusion}
MDPipe makes a significant novel contribution to ocular disease diagnosis by harnessing LLMs to address the limitations of current diagnostic methods. We integrate a visual translator that quantifies the visual data and clinician expertise for robust reasoning. As evidenced by  benchmarking and the user study, MDPipe demonstrates a significant improvement in accurate OSD diagnosis with clinically relevant rationale. For future work, expanding the dataset of real-life clinician diagnoses and clinically verified variable scales promises to further refine the robustness and depth of LLM-generated rationales.
\begin{credits}
\subsubsection{\ackname} This work is supported by NIH (R21EY033881); UC Berkeley, Clinical Research Center Unrestricted Fund; and the Roberta J. Smith Research Fund.

\subsubsection{\discintname}
The authors have no competing interests to declare that are
relevant to the content of this article.
\end{credits}
%
%
%
%
\bibliographystyle{splncs04}
\bibliography{Paper-0298}

\clearpage

\appendix

\section*{\Large Appendix} 
In Section~\ref{sec:appendix-implement}, we first provide additional implementation details for 1) LLM fine-tuning, 2) visual translator, and 3) clinical report summarizer. 
Then in Section~\ref{sec:appendix-discuss}, we give additional discussion about our approach and dataset. In Section~\ref{sec:appendix-template}, we provide the template (to GPT-4) for LLM-based summarizer. 

\section{Implementation Details}
\label{sec:appendix-implement}

\noindent\textbf{LLM Fine-Tuning.} We leverage the TRL-Transformer Reinforcement Learning GitHub repo to fine-tune Llama2. We utilized the TRL library incorporating 4-bit QLoRA for efficient training. Fine-tuning was conducted with a batch size of 4 per device with maximum 10,000 steps. Learning rate was 2e-4 with a constant scheduler and maintained a maximum sequence length of 512. All other parameters align with those specified in the TRL GitHub repo.

\noindent\textbf{Visual Translator.} Our implementation is based on the Wizaron/instance-segmentation-pytorch GitHub repo. We utilized a ResNet50 backbone for our instance segmentation network, training separate models for lower (553 images) and upper (486 images) eyelids on CRC data annotated with gland masks. Each model was trained on 256x256 resized images for 300 epochs using a batch size of 8 and a learning rate of 1.0, employing the Adadelta optimizer with a weight decay of 1e-3. Other parameters align with those specified in the GitHub repo.

\noindent\textbf{LLM-Based Clinical Report Summarizer.} The summarizer uses the GPT-4 API in the Erol444/gpt4-openai-api GitHub repo, with a unique seed to control uncertainty. We input raw morphology data along with a structured task template (Section~\ref{sec:appendix-template}) to produce clinical reports via GPT-4.

\section{Discussions}
\label{sec:appendix-discuss}

\noindent\textbf{Does MGD Imply DE?} The presence of Meibomian Gland Dysfunction (MGD) does not necessarily imply evaporative Dry Eye (DE). While MGD is a common factor in evaporative DE, it is not always the case (Galor, 2014). MGD, DE, and blepharitis are distinct conditions with overlapping symptoms. In our model, we use independent labels for these conditions based on the TFOS 2017 DEWS II Definition and Classification Report (Craig et al., 2017). DE is marked by tear film instability, ocular surface damage, and symptoms. MGD involves ductal stenosis and gland secretion quality, while blepharitis is identified by eyelid margin inflammation, debris, and collarettes. Differentiating these conditions in our model enhances diagnostic accuracy and treatment efficacy.

\noindent\textbf{Is our Dataset Diverse Enough?} It is important to note that our datasets come from diverse study populations. Data from the CRC and DREAM (a major clinical trial with 11 meibography sites across the US) are combined. Our distributions closely align with US Census statistics for age, sex, and race. Additionally, our dataset encompasses a wide range of disease severities. Efforts are ongoing to obtain additional data from much younger and older populations, male subjects, and individuals of African ethnicity.

\section{Template for LLM-based Summarizer}
\label{sec:appendix-template}

\begin{bkg_template}
\sloppy{\texttt{\bf{Task Description:}}} \\
\sloppy{\texttt{You are an intelligent medical summary generator. Your task is to generate a clinical report summary for the raw clinical metadata mentioned in the caption.
I will provide you with medical data obtained from meibography images of patients and generate concise summaries. Please generate a human readable summary with Q\&A format by setting subject's demography, and MG morphology as the Question while the ocular surface disease as the Answer.}} \\\\
\sloppy{\texttt{\bf{Supporting Examples:}}} \\
\sloppy{\texttt{Here's an example:
 \{"42\_2\_R": \{"gender": "Male", "age": 30, "race": "Asian", "TMH": 0.28, "NIKBUT": 12.33, "MG\_Morph": \{"avg\_length": 4.878048, "avg\_width": 0.463366, "avg\_contrast": 13.573304, "avg\_tortuosity": 0.277598\}, "Dry Eye": "Yes", "Meibomian Gland Dysfunction": "No", "Blepharitis": "Yes"\}\}}}\\
\sloppy{\texttt{You should output something like: \\
\#\#\#Human: Subject 42\_2\_R and right eye. The person is a male with an age of 30, and the race is Asian. The Tear Meniscus Height (TMH) is 0.28mm, The Non-Invasive Keratograph Tear Breakup Time (NIKBUT) is 12.33 sec. The meibomian gland morphology has average length of 4.87mm, average width of 0.46mm, avg contrast is 13.57, and average tortuosity is 0.28.\\
\#\#\#Assistant: The Dry Eye (DE) condition for this subject is Yes, and The Meibomian Gland Dysfunction (MGD) is No, and the Blepharitis is also Yes.}}\\\\
\sloppy{\texttt{This is the patient 42\_2\_R, 2 means OS2 category, 42 is patient ID, R is right eye. MG\_Morph means meibomian gland morphology features. Please remove values after second decimal place. Please write in one paragraph as a clinical report summary. This summary could be an input data to fine tune large language model.}}\\

\sloppy{\texttt{\bf{Prompting the Clinical Metadata}}} \\
\sloppy{\texttt{Could you give a clinical report summary of the data? \bf{\{your input metadata\}}}}

\end{bkg_template}

\end{document}